# Modeling Individual Differences in Game Behavior using HMM

Sara Bunian[1,4], Alessandro Canossa[5], Randy Colvin[2,4], Magy Seif El-Nasr[3,4]

[1] Electrical and Computer Engineering Department [2] Department of Psychology [3] College of Computer and Information Science

[4] Northeastern University, Boston [5] Ubisoft, Sweden

banian.s@husky.edu.neu, aleandait@gmail.com, r.colvin@neu.edu, m.seifel-nasr@neu.edu

**Abstract**

Player modeling is an important concept that has gained much attention in game research due to its utility in developing adaptive techniques to target better designs for engagement and retention. Previous work has explored modeling individual differences using machine learning algorithms performed on aggregated game actions. However, players' individual differences may be better manifested through sequential patterns of the in-game player's actions. While few works have explored sequential analysis of player data, none have explored the use of Hidden Markov Models (HMM) to model individual differences, which is the topic of this paper. In particular, we developed a modeling approach using data collected from players playing a Role-Playing Game (RPG). Our proposed approach is two fold: 1. We present a Hidden Markov Model (HMM) of player in-game behaviors to model individual differences, and 2. using the output of the HMM, we generate behavioral features used to classify real world players' characteristics, including game expertise and the big five personality traits. Our results show predictive power for some of personality traits, such as game expertise and conscientiousness, but the most influential factor was game expertise.

## Introduction

While games have been developed for many years for many sectors and applications, including entertainment, training, health, and education (Goldman 2010), developing such games is not an easy process and most suffer from low retention rates (Farago 2011 and 2012). The need to develop more successful games spur the need for work to investigate new approaches to enhance game design and improve player experience. One approach is to develop player modeling systems that adapt the design (Yannakakis et al. 2013)

Player modeling has gained considerable momentum recently from both industry and academic research. It is defined as the process of developing computational models to describe player behaviors in a game environment (Yannakakis et al. 2013). Many commercial games have adopted its use for different purposes, such as Dynamic Difficulty Adjustments in Left 4 Dead (Nagle, Wolf, and Riener 2016), designing new content updates for League of Legends (Lewis and Dill 2015). Within the research community, player modeling has been used for churn prediction (Mahlmann et al. 2010), dynamic difficulty adjustment (Missura and Gärtner 2009), goal recognition (Min et al. 2016), and understanding strategies (Weber and Mateas 2009). However, little work has explored modeling players' individual difference, such as personality or expertise, using in-game behaviors, while some HCI work has examined the role of individual differences on engagement and design (Halko et al. 2010; Tuten et al. 2001; Gao et al. 2013). We thus argue that modeling individual differences will have benefits in terms of player enjoyment, inclusive designs, adaptability, and user research (Orji et al. 2017).

Individual differences and preferences are usually defined using the Five Factor Model (FFM) (Goldberg 1990). Generally, the model consists of five traits, which are defined as openness to experience, conscientiousness, extraversion, agreeableness, and neuroticism, each relating to different attributes, that account for diverse aspects of personality. To understand such constructs and their manifestation in behavior, it is important to account for sequential behavioral patterns. However, current work on player modeling use aggregate features, such as completion time, and enemies killed (Canossa et al. 2015; Erfani et al. 2010; Yee et al. 2011), which is limited. In this paper, we analyze sequential player actions to model individual differences and predict it using in-game behaviors.

Out of the many game genres available, role-player games (RPG) provide a game world where players are provided with breadth of activities. These activities hold within challenges, puzzles, and dilemmas, which make RPGs a rich and interesting environment to analyze player behavior and model individual difference.

In this paper, we hypothesize that following player action choices over time in an RPG game with various affordances might reveal hidden individual behavioral differences. We focus on modeling player behavior via game data logs where players interact in an RPG game, called VPAL (Virtual Personality Assessment Lab). In addition to game play logs, we also collect personality data through multiple sources including self-report questionnaires and behavioral measures, and game expertise through a self-report measure developed previously (Joorabchi and Seif El-Nasr 2011). Using the game log data, we develop a Hidden Markov Model (HMM), where the collection of players' drives (e.g. strategies, tactics, and actions) at any moment is represented as a hidden state that influences how a player chooses to behave in the virtual world. The model processes the sequence of player's actions to define a sequence of hidden states that best describes the behavioral changes that occurred during play. The output of the model, represented as the frequencies of the player's behavioral states, is used to generate new behavioral features. These features are fed into a logistic regression classification model to predict real world characteristics, specifically game expertise and the big five personality traits.

The contribution of this work is two fold: 1) it presents a novel approach to develop a sequential probabilistic model to describe player behaviors within an RPG game using game data logs to model individual differences, and 2) it presents results to show the predictive power of game data in uncovering personality and game expertise. To our knowledge, this is the first work that utilizes sequential probabilistic algorithm of in-game player behaviors to model individual differences, in terms of personality and game expertise.

The remainder of this paper is organized as follows: Section 2 reviews related work. Section 3 describes the dataset and the testbed game used to validate the method. The design and methodology of the proposed model is described in Section 4. The results of the model evaluation are presented in Section 5. Section 6 offers concluding remarks.

## Related Work

Player modeling is an active field of game research that has recently gained momentum for its utility to create models for difficulty adjustments, churn prediction, strategy prediction, adapting player experience, playtesting, game authoring, and personalized content generation. For example, a series of classification algorithms, such as NNge, M5, and linear regression, are presented in (Weber and Mateas 2009) to predict player strategies in StarCraft. Using in-game play data Mahlmann et al. (2010) trained logistic regression, Bayes Network, Decision Trees, and Support Vector Machine, to predict churn for Tomb Raider: Underworld (Eidos Interactive 2008). Missura and Gärtner (2009) explored player classification to dynamically adjust the difficulty of a shooter game by exploring the use of K-means clustering and support vector machines.

In addition to the use of aggregate data and classification methods, Harrison and Roberts (2011) designed a predictive model of user behavior in World of Warcraft, based only on the previous sequence of observations that occurred in similar situations. Specifically, they utilize previous player histories to create groups of similar actions that are then used to predict the user behavior and action. Valls-Vargas et al. (2015) proposed an approach that combines episodic segmentation of gameplay traces and sequential machine learning to dynamically predict play styles.

Of particular interest to us are previous works on modeling of players' individual differences. Commonly, the basic approach is to characterize player profiles based on gaming archetypes. A seminal work in this area is the work of Bartle (Bartle 1996), who proposed a player model categorizing players into four types: Achievers, Socializers, Explorers, and Killers. Using Myers-Briggs personality indicator (Myers-Briggs 1962), Bateman and Boon (2005) created another model that identifies players into four types: Conqueror, Manager, Wanderer, and Participant, which they later expanded to seven categories: Seeker, Survivor, Daredevil, Mastermind, Conqueror, Socializer and Achiever, and loosely. Further, Canossa (2009) developed player persona using clustering algorithms and in-game player data.

In addition to player archetypes, researchers constructed player models using psychological personality theories. For example, using Reiss's individual differences model (Reiss, 2008), Canossa et al. (2013) explored the psychological motivations governing players of Minecraft. Specifically, they collected in-game behaviors from 546 players, out of these they collected personality data using Reiss' test from 90 of them. Results of correlation analysis demonstrated that motivational factors, particularly: honor, romance, independence and acceptance, were highly correlated to in-game behaviors. In addition, researchers have also used the FFM (Costa and McCrae, 1992) to define player behaviors (Canossa et al., 2015; Chen et al., 2015; Van Lankveld, Schreurs, and Spronck, 2009; Van Lankveld et al., 2011; Yee et al., 2011). For example, Yee et al. (2011) investigated correlation between in-game behaviors and FFM for World of Warcraft. Their findings demonstrated significant correlations for each of the five personality traits, although no coefficients exceeded ±0.17. To form a more precise description of personality, Canossa et al. (2015) explored the relationship between the FFM facets and in-game behaviors in various game contexts in the VPAL game. Their results identified much stronger correlation when individual personality facets and game context is used. On the other hand, an Lankveld et al. (2009; 2011) focused on specifically one personality trait where they explored the correlation between Extraversion and behaviors in a customized scenario of Neverwinter Nights. They found significant correlations for extraversion in the range ±0.32.

In addition to correlating personality measures with game behavior, Chen et al. (2015) recently utilized sequence pattern mining to further understand the impact of individual differences on game behaviors. They developed a model, using closed sequential pattern mining and logistic regression, that uses gameplay action sequences in an RPG game to predict gender, game expertise and the FFM personality traits. Results showed the possibility of extracting behavioral features to understand such differences, with game expertise being the most effective factor (with reported error rate of 0.25, accuracy of logistic regression was not reported).

Researchers have also investigated modeling player profiles through understanding the relationship between in-game behaviors and individual differences in terms of gender and game expertise (Erfani et al. 2010; Milam, Bartram, and Seif El-Nasr 2012). Milam et al. (2012) identified the significant effect of game expertise on critical metrics of a railed-shooter game, such as accuracy and level time. Erfani et al. (2010) investigated the influence of age, gender and game expertise on game performance. They conducted an experiment with 60 kids on 3 games and concluded that these factors have an impact on player performance.

Our approach to player modeling can be viewed as similar to the work presented in (Chen et al. 2015) but different in terms of the machine learning algorithm used. Specifically, we utilized an HMM model to model behavioral patterns over time to predict individual differences. To our knowledge, this is the first work that uses HMM to investigate individual differences.

## Game and Data

The test game used to validate our model is VPAL (Virtual Personality Assessment Lab) game – a 3D role-playing open world game designed using the Fallout New Vegas 3D engine, screenshots from the game are shown in Figure 1. VPAL tells the story of a generic town that has been taken over by bikers. In an attempt to discover the town and save it, the player controls a character through numerous possible interactions, including exploring different areas such as hotel or mine, collecting different objects and weapons, conversing with Non-Playing characters (NPCs), engaging in combats, and performing quests.

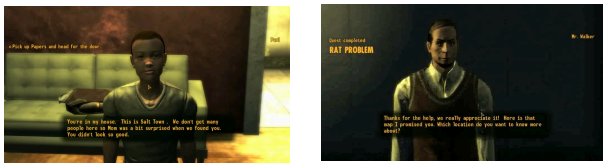

*Figure 1: Screenshots from VPAL the game*

For this paper, we will focus on the first level, which takes place in the Intro House – a small house with several rooms where players can perform a wide range of behaviors, e.g., engage with three NPCs and perform up to three quests. They game starts with accepting the first mandatory quest from Paul.

### Data

Using VPAL, we collected game logs of player behavior from 66 participants, who were asked to play the game for 60 minutes in the lab. The game logs include all player actions time stamped, including movement information, quests details, dialogue choices, object and NPC interactions (talking, fighting, killing). A valid action is represented as a set of tokens that starts with the action type and is supported by additional information, such as the location of action.

In addition to in-game player behavior data, we also collected self report data of the players' personality and game expertise. For expertise, we used a previously developed and validated survey (Joorabchi and Seif El-Nasr 2011). For personality scores, we used the NEO-PI-R (Costa and McCrae 1992).

## Our Model

### Preprocessing of in-game data

To create the model, we first preprocessed, cleaned, and parsed the data to produce abstracted actions for each player. Due to the limited activities performed in the first level of the game, we only processed abstracted actions which are listed in Table 1. We selected actions for analysis that had an occurrence rate of 10% or greater because actions performed by few players would not help discriminate between groups.

| Abbreviation | Action |
|---|---|
| SQ | Start Quest |
| CQ | Complete Quest |
| D | Normal Dialogue |
| DT | Dialogues with soliciting behavior |
| DR | Dialogues with Rude behavior |
| A | Random Attack (Unmotivated, Friendly NPC) |
| AQ | Quest Related Attack |
| I | Interaction with environment's Objects |
| IN | Interaction with NPC |
| U | Use Weapon/Item |
| E | Equip Weapon/Item |
| K | Kill |
| L | Loot Item/Player |

*Table 1: List of in-game actions*

### HMM Model

An HMM is a generative probabilistic model of time series data, mainly used for analyzing and recovering a sequence of internal hidden states generated by a sequence of output

observations (Rabiner and Juang 1986). An HMM model is characterized by parameters represented by Eq. (1):

$$\lambda = (S_t, O_t, A, B, \pi) \quad (1)$$

Where $S_t$ represents the finite set of the hidden states; $O_t$ is the finite set of observed outputs; $A$ is the state transition probability matrix that defines the probability from going from one state to another; $B$ is the emission matrix (i.e. observation probability matrix) that defines how each observation contributes to each state; and $\pi$ is the initial state probability matrix that represents the probability of being in a given state at the start of sequence.

HMM is one of the most popular probabilistic models that have been applied intensively to investigate hidden behavioral patterns in various fields (e.g., Jeong et al. 2008, Boussemart et al. 2009, Carola, Mirabeau, and Gross 2011, Tang et al. 2016). To analyze VPAL behaviors, we created a mapping of the HMM parameters to the actual game scenario, where the hidden states ($S_t$) represent the player's time-dependent strategy of their behaviors and the observed output ($O_t$) represent abstractions of the observed low-level actions executed by the player (see Table 1).

**Determining the number of hidden states**

For HMM, one of the important aspects of evaluating the robustness of the model is to correctly determine the number of hidden states ($S$) to be used. To decide on the best choice, we ran the *Baum-Welch* algorithm on the data for several $S$s then calculated the *Bayesian Information Criterion (BIC)* for each model. The best model is the one that minimizes the BIC score according to Eq. (2):

$$BIC = -2\, logLikelihood + D \log P \quad (2)$$

Where $P$ is proportional to the size of the data, $D$ is the number of parameters for the model, and *logLikelihood* is the likelihood of the data given the HMM, as computed through the forward-backward procedure. $D$ is calculated using Eq. (3):

$$D = N - 1 + N(M - 1) + N(N - 1) \quad (3)$$

Where $N$ is the number of states and $M$ is the number distinct player actions.

After training different models of different sizes, we chose $S$=5 due to its optimal BIC (see Figure 2).

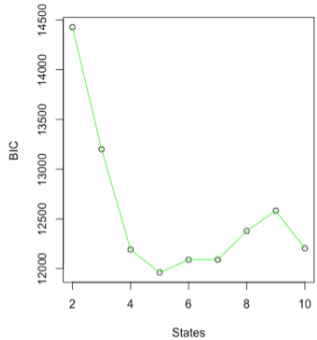

*Figure 2: BIC Score for varying model sizes*

**The algorithm and Modeling Process**

To analyze player behaviors using HMM, we adopted the model presented in (Carola, Mirabeau, and Gross 2011), which is depicted in Figure 3. We used hidden states to model the players' cognitive states and strategies that are not directly observed. We, thus, started with training the HMM model on all the players using the *Baum-Welch algorithm* (Stratonovich 1960), which is an Expectation-Maximization (EM) algorithm that learns the best parameters from the data. The algorithm iteratively adjusts the parameters of the model to maximize the likelihood that the sequence of observed data was generated by the HMM. Due to absence of having a priori of what the expected hidden states should be, we used the EM algorithm in an unsupervised mode, which utilizes Bayesian inference to automatically infer the optimal parameters of the model (Boussemart et al. 2009). The algorithm uses a gradient search in the parameter space to optimize the likelihood.

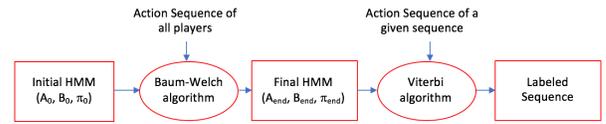

*Figure 3: Structure of proposed model*

Once we populated the model, we used the Viterbi algorithm (Forney 1973) to compute the hidden sequence where player's style changed from the observed sequence. Specifically, we used the Viterbi Algorithm to label the sequence of observed actions for each player. Using this algorithm, we uncovered the optimal path of the most probable behavioral sequence that caused the observations.

**Predictive Model**

Based on the output of the HMM, consisting of frequency of being in a state, we predicted six categories of individual differences, which are game expertise and the five personality traits. Specifically, we model the probability P(Category|State Frequencies). For each category, we applied a binarization process, where a player is considered to have a *high* degree of one trait if the score is above the mean, and a *low* degree if his/her trait score is below the mean.

The input feature vector is defined as:

$$\phi(x) = (x_1\ x_2\ x_3\ x_4\ x_5) \quad (4)$$

where each feature $x_i$ represents the frequency of being in each of the five hidden states of the HMM. Thus, we create 6 classification models for each category and the $P(y=1|X)$ represents a high score players in this trait.

# Results

## The Model

Figure 4 shows the overall graphical representation of the final HMM in terms of the hidden states (S1 through S5) and the coefficients of the derived emission matrix. Each hidden state is comprised of a set of dominant activity(ies) that are used to set the state label. Some states are governed by a single dominant activity, whereas others are governed by multiple activities. Thus, states are described as the players' behaviors. In this case, state 1 corresponds to behaviors consistent with Exploring (E), including Interaction with objects (I), looting items (L), and completing quests (CQ), while state 2 is mostly described by interacting with NPCs (IN) thus engaging with characters (N). State 3 corresponds to behaviors consistent with conversing with NPCs (D), i.e. being social (S). State 4 is governed by behaviors related to completing quests including using weapons (U) and quest related attacks (AQ), therefore has been labeled Achieving (A). State 5 however is mostly dominated by the attacking activity (A) which corresponds to hostility behavior (V).

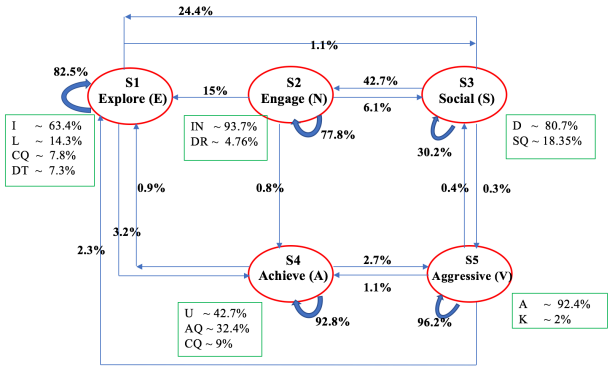

*Figure 4: General Structure of HMM*

To illustrate the process of HMM, see table 2. The table presents sequences of observed actions for two different players and the corresponding HMM states. It's clear, by looking at the HMM states that both players started at the same social state conversing with NPC, Paul, about the mandatory quest. However, afterwards each player exhibited a different pathway of behaviors. For example, player 1 spent significantly more time in the exploring and engaging states while never going to the aggressive state. Also, the player managed to achieve and complete the quest in a systematic manner, starting with collecting necessary information from Mr. Walker, attacking the rat and killing it, which could reveal certain deterministic behaviors. However, player 2 followed a different approach in completing the level. It's evident that he spent much of his time in the aggressiveness state engaging in unmotivated attacks to friendly NPCs. Also, the player never attempted to complete the quest (as he never went to the A state). By looking at the dialogue choices at the end of the sequences, we can notice how player 1 was trying to collect information from the NPC about the surrounding environment and the available quests, wherein player 2 engaged in random conversations before deicing to leave the house. Comparing the behavioral findings from the sequences revealed similarities to some of the personality scores; player 1 scored very high in extroversion, openness, and conscientiousness while player 2 scored very low in extroversion.

| **Sequence #1** | |
|---|---|
| Actions | SQ D D D CQ I IN D SQ D I I IN D D D SQ D D D D D D D D D I I I I I L I I AQ AQ AQ AQ AQ CQ K L L L L L L L I IN CQ DT DT DT I CQ |
| HMM States | N S S S E E N S S S E E N S S S S S S S S S S S S E E E E E E E A A A A A A A A A A A A A E N E E E E E E |
| State Frequencies | S1:19, S2: 4, S3:19, S4: 14, S5:0 |
| **Sequence #2** | |
| Actions | SQ D D IN A I I IN A A A A A A A CQ A A A I IN D D DR IN D D D SQ D D D D D D D D D D I I I I |
| HMM States | N S S N V E E N V V V V V V V V V V V E N S S N N S S S S S S S S S S S S E E E |
| State Frequencies | S1:7, S2:6, S3:17, S4:0, S5: 13 |

*Table 2: Sequence of actions for two different players*

## Behavioral Features to Discriminate Individual Differences: Descriptive Statistics

We labeled all participants and calculated frequencies of states for all players. Results of the differences between frequencies of states per individual difference parameter is shown in Figure 5. We are only showing Game Expertise, Extroversion, and Conscientiousness with mean of state frequencies displayed. We used ANOVA to measure the significance of variance among the top/low 15 participants of each personality in terms of each behavioral state. We found no significant results for the categories Openness, Agreeableness, and Neuroticism ($p > 0.05$). This can be attributed to the limited environment of the IntroHouse and the context of the available interactions within it. However, we found significant differences for some of the states in game expertise, extraversion, and Conscientiousness.

Results showed that for game expertise there are significant differences between participants in S3 (S) and S4 (A) with ($p=0.0028$ and $p =0.0104$, respectively), where players with high expertise spent much time achieving and completing the given quests (S4), compared to those with low expertise players. Players with low expertise spent more time conversing with NPCs (S3). It's also clear from the figure that experienced players almost never exhibit aggressive behaviors (S5), wherein low experienced players seem to spend more time engaged in aggressive behaviors. For Extraversion, we found significant effects in state (S5) with (p= 0.0172), which indicates that high extroverts exhibit very little aggressive behaviors compared to low extroverts. Also, for Conscientiousness, S3 frequency was significant (p= 0.00629), which reveals that participants who are more conscientious tend to converse significantly more with NPCs (S3) than participants low on conscientiousness. We could not show significance for any other state frequencies.

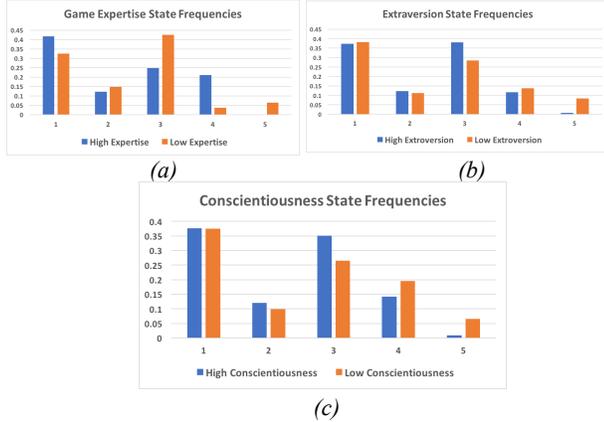

*Figure 5: Average State Frequencies for categories with significant variance (a) Game expertise ((b) Extraversion, (c) Conscientiousness*

### Prediction of Individual Differences using Behavioral Features from HMM

To evaluate the effectiveness of our behavioral HMM, we developed a predictive model for each category based on the newly generated behavioral features, which correspond to the state frequencies. We wanted to measure how important these features in predicting player behavior. To do so, we trained a logistic regression (LR) model based on the composition of the newly generated behavioral features for the 66 participants. Considering the small dataset, we used only 3-fold cross validation. The prediction accuracy for each category is presented in Table 3. It can be seen that the composition of all the state frequencies contribute the most to game expertise and conscientiousness category as the LR provided an accuracy of 70.13% and 59.1%, respectively (shown by * in the table). However, the amount of time spent in each state was not predictive.

To see if our assumption of using sequential patterns than aggregate data is valid, we compared the predictive accuracy of the newly generated behavioral features to the aggregated features (e.g. Number of dialogues, Number of quests completed, etc.). The same procedure for training logistic regression was applied on the data, except that the features represent the composition of the aggregates of the 13 behaviors extracted in Table 1. The predictive accuracy for each of the personality categories is shown in Table 3 (where accuracies that are higher for HMM is bolded). As can be seen, except for openness and conscientiousness, overall the accuracy of the other features is higher in the HMM especially for game expertise. Thus, by using HMM features to form a description of personality categories, we were able to increase the power of personality to explain behaviors. This can be interpreted that, for some categories, HMM is able to precisely model behaviors to better identify individual differences. The most dominant factor in our analysis is game expertise, this is compatible with findings reported in (Chen et al. 2015; Joorabchi and Seif El-Nasr 2011; Milam, Bartram, and Seif El-Nasr 2012).

| Category | Prediction Accuracy | |
|---|---|---|
| | *Model 1: HMM Generated Features* | *Model 2: Aggregate Statistics* |
| Game Expertise | **70.13%*** | 60.9% |
| Extraversion | **57.6%** | 55.1% |
| Openness | 54.1% | 57.89% |
| Conscientiousness | 59.1%* | 60.1% |
| Agreeableness | **57.7%** | 42.1% |
| Neuroticism | **57.9%** | 46.8% |

*Table 3: Predictive accuracy of the logistic regression for two different models: using HMM states and aggregated statistics*

## Conclusion

In this project, we used the data from an RPG game called VPAL game to analyze and model individual differences exhibited as patterns of in-game behaviors. The observed sequence of actions performed in the game was used as input to an HMM Model to estimate the behavioral changes of player style. Generally, the model clustered behaviors and defined higher level behavioral states: explore, engage, social, achieve, aggressive. The *Viterbi* algorithm utilized the general model to find the optimal path of sequences for each player. To compare the output of the model with the collected personality scores, we generated new behavioral features using state frequencies and compared the top/low participant of each personality trait. Results of analysis of variance indicated noticeable in state frequencies used by game experts and non-experts as well as extroverts vs. introverts, and Conscientiousness participants, where experts engaged more in quest activities and less in aggressive activities than non-experts, extroverts exhibited more aggressive behaviors than introverts, and conscientious participants engages in more social activities with NPCs. HMM showed higher accuracy for predicting game expertise and conscientiousness, and in general displayed better accuracy than using aggregate data. We believe our work will mark a good contribution to player modeling and game analytics, and can show the impact of individual differences on player behaviors.